\documentclass[runningheads]{llncs}
\usepackage{graphicx}

\usepackage[utf8]{inputenc} 
\usepackage[T1]{fontenc}    
\usepackage{hyperref}       
\usepackage{url}            
\usepackage{booktabs}       
\usepackage{amsfonts}       
\usepackage{nicefrac}       
\usepackage{microtype}      
\usepackage{xcolor}         
\usepackage{multirow}
\usepackage{amsmath}

\usepackage{algorithm}
\usepackage{algorithmic}

\begin{document}
\title{Principal Components Analysis based frameworks for efficient missing data imputation algorithms}

%



\author{Thu Nguyen\inst{1}
\and
Hoang Thien Ly\inst{2} \and 
Hugo Michael \\ Alexander Riegler\inst{1}
\and Pål Halvorsen\inst{1}
\and Hugo L. Hammer\inst{1}}
\authorrunning{T. Nguyen et al.}
%
\institute{SimulaMet, Norway
\and
Warsaw University of Technology, Poland}


\maketitle

\begin{abstract}
Missing data is a commonly occurring problem in practice. Many imputation methods have been developed to fill in the missing entries. However, not all of them can scale to high-dimensional data, especially the multiple imputation techniques. Meanwhile, the data nowadays tends toward high-dimensional.   
Therefore, we propose \textit{Principal Component Analysis Imputation} (PCAI), a simple but versatile framework based on Principal Component Analysis (PCA) to speed up the imputation process and alleviate memory issues of many available imputation techniques while maintaining good imputation quality. In addition, the frameworks can be used even when some or all of the missing features are categorical or when the number of missing features is large. We also analyze the effect of using different formulations of PCA on the technique.
Next, we introduce  \textit{PCA Imputation - Classification} (PIC), an application of PCAI for classification problems with some adjustments.
Experiments on various scenarios show that PCAI and PIC can work with various imputation algorithms, including state-of-the-art ones, and improve the imputation speed significantly while achieving competitive mean square error/classification accuracy compared to imputing directly on the missing data.

\end{abstract}

 \section{Introduction}\label{sec-intro}

Despite recent efforts in directly handling missing data \cite{nguyen4260235pmf,NGUYEN20211,lipton2016modeling}, imputation approaches \cite{buuren2010mice,yoon2018gain} remain widely used. This is because directly handling missing data can be complicated and usually are developed for specific target problems or models. Meanwhile, imputation can be more versatile as it makes the data \emph{complete}, i.e., no longer have any missing values. Therefore, it is easier to continue with other preprocessing steps, analysis, and data visualizations. 
Meanwhile, directly handling missing data strategies are often more complicated and not that readily available.

Many techniques have been developed for missing data imputation \cite{buuren2010mice,yoon2018gain,vu2023conditional}, 
most  of them are computationally expensive for big datasets. For example, experiments in \cite{NGUYEN2022108082} show that under their experiment settings, for Fashion MNIST
, a dataset of 70,000 samples and 784 features, MICE \cite{buuren2010mice} and missForest \cite{stekhoven2012missforest} are unable to finish the imputation process within three hours for a missing rate (the ratio between the number of missing entries versus the total number entries in the dataset) of 20\%. 
Since datasets nowadays are trending towards larger sizes,  
with hundreds of thousands of features \cite{guyon2007competitive}, it is crucial to speed up the available imputation techniques. Taking into account resource consumption and availability such speed up cannot be achieved by only providing more and better hardware but by the development of new methods.

To achieve this goal, this work introduces two novel frameworks based on Principal Component Analysis (PCA) to speed up the imputation process of many available techniques or the imputation-classification process for missing data classification problems. The first framework, \textbf{PCA Imputation (PCAI)} is proposed to speed up the imputation speed by partitioning the data into the fully observed features partition and the partition of features with missing data. After that, the imputation of the missing part is performed based on the union of the PCA - reduced version of the fully observed part and the missing part. Interestingly, it turns out that the method has a great potential to aid the performance of methods that rely on many parameters, such as deep learning imputation techniques. 
Meanwhile, the second one, \textbf{PCA Imputation - Classification (PIC)} is proposed to deal with the missing data classification problems where dimension reduction is desirable in advance of the model training step. PIC is based on PCAI with some modifications. Note that these frameworks are different from the methods developed for principal component analysis under missing data presented in \cite{grung1998missing,ilin2010practical}, which are about how to conduct PCA when the data contains missing values. 

In summary, the contributions of this article are: (i) we introduce \textbf{PCAI} to improve the imputation speed of many available imputation techniques; (ii) we introduce \textbf{PIC} to deal with missing data classification problems where dimension reduction is desirable; (iii) we analyze the potential strength and drawbacks of these approaches; and  (iv) we illustrate via experiments that our frameworks can work with various imputation strategies while achieve comparable or even lower mean square error/higher classification accuracies compared to the corresponding original approaches, and alleviate the memory issue in some approaches.


The rest of the paper is organized as follows. In Section \ref{prelim}, we review two popular formulations of PCA. Next, in Section \ref{sec-pcai}, Section \ref{sec-pic}, and Section \ref{sec-related}, we introduce our novel PCAI and PIC frameworks, and study related works, respectively. After that, in Section \ref{sec-exper}, we demonstrate their capabilities via experiments on various datasets. The paper ends with conclusions, remarks, and future works in Section \ref{sec-concl}. 

\section{Preliminaries}\label{prelim}
Let $\mathbf{X} = [x_{ij}]$ where $i = 1,...,n; j = 1,...,p$ be an input data matrix of $n$ samples, $p$ features. In addition, assume that the features are centered and scaled. We review two popular formulations of PCA, which we refer to as \textbf{PCA-form1} and \textbf{PCA-form2}.
\paragraph{PCA based on covariance matrix (\textbf{PCA-form1}).} 
Let $\mathbf{\Sigma }$ be the covariance matrix of $\mathbf{X}$. Next, let $(\lambda_1,\mathbf{v}_1), ...,(\lambda_p,\mathbf{v}_p)$ be the sorted eigenvalue-eigenvector pairs of $\mathbf{\Sigma}$ such that $\lambda_1 \ge \lambda_2 \ge ...\ge \lambda_p\ge 0$. Suppose that we choose the first $r$ pairs for dimension reduction. Then the amount of variance explained by these $r$ pairs is
\begin{equation}
    \frac{\lambda_1+\lambda_2+...+\lambda_r}{\lambda_1+\lambda_2+...+\lambda_p}
\end{equation}
Next, let $\mathbf{V} = [\mathbf{v}_1, \mathbf{v}_2,..., \mathbf{v}_r].$
Then, the dimension reduced version of $\mathbf{X}$ is $\mathbf{X}\mathbf{V}$.

\paragraph{PCA based on the input matrix $\mathbf{X}$ (\textbf{PCA-form2}).} 
The solution of PCA can also be produced based on the singular value decomposition of $\mathbf{X}$:
\begin{equation}
    \mathbf{X} = \mathbf{U}\mathbf{D}\mathbf{W}^T   
\end{equation}
where $\mathbf{U}$ is an $n\times p$ orthogonal matrix, $\mathbf{W}$ is a $p\times p$ orthogonal matrix, and $\mathbf{D}$ is a $p\times p$ diagonal matrix whose diagonal elements are $d_1 \ge d_2 \ge ...\ge d_p\ge 0$. Suppose that $r$ eigenvalues are used, then the projection matrix is $\mathbf{V}=\mathbf{W}_r$ where $\mathbf{W}_r$ consists of the first $r$ columns of $\mathbf{W}$.  Then the dimension reduced version of $\mathbf{X}$ is also $\mathbf{X}\mathbf{V}$.
\section{PCA Imputation (PCAI)}\label{sec-pcai}


To start with some notations, let $pca(A)$ be a function of a data matrix $A$. 
The function returns $(\mathcal{R}_A, V)$ where $\mathcal{R}_A$ is the PCA-reduced version of $A$, and $V$ is the projection matrix where the $i^{th}$ column of $V$ is the eigenvector corresponding to the $i^{th}$ largest eigenvalue. In addition, denote by $[\mathcal{A}, \mathcal{B}]$ the columnwise concatenation of two data partition $\mathcal{A}$ and $\mathcal{B}$ of relevant sizes. Next, suppose that we have a dataset $\mathcal{D} = [\mathcal{F}, \mathcal{M}]$ where $\mathcal{F}$ consists of data from fully observed features and $\mathcal{M}$ consists of data from features with missing values.

The framework is as depicted in Algorithm \ref{alg-pcai}. We first conduct dimension reduction on the fully observed partition $\mathcal{F}$, which produces a reduced version $\mathcal{R}$ of $\mathcal{F}$. Then, the imputation of $\mathcal{M}$ is done on the set $[\mathcal{R}, \mathcal{M}]$ instead of $\mathcal{D} = [\mathcal{F}, \mathcal{M}]$ as how imputations are usually done (i.e., impute directly on the original missing data). In conducting dimension reduction, we expect to reduce the dimension of the fully observed partition so that the imputation of $\mathcal{M}$ can be faster. 


\begin{algorithm}
\caption{PCAI framework}\label{alg-pcai}
\begin{algorithmic}
\scriptsize
\REQUIRE 
\STATE - $\mathcal{D} = [\mathcal{F}, \mathcal{M}]$ where $\mathcal{F}$ is the fully observed partition and $\mathcal{M}$ is the partition with missing values
\STATE - Imputer $I$ , PCA algorithm $pca$

\hspace{-.4cm}\textbf{Procedure:}

\STATE $(\mathcal{R}, V) \leftarrow pca(\mathcal{F})$
\STATE $\mathcal{M}'\leftarrow $ the imputed version of $\mathcal{M}$ based on $[\mathcal{R}, \mathcal{M}]$ 
\RETURN Imputed version $\mathcal{M}'$ of $\mathcal{M}$
\end{algorithmic}
\end{algorithm}

       


       

For the choice of the PCA formulation, note that if the number of samples is larger than the number of features in $\mathcal{F}$, then the size of the covariance matrix is smaller than the size of $\mathcal{F}$. Therefore, one may expect using the formulation of PCA based on the covariance matrix (PCA-form1), to be faster. Meanwhile, if the number of features in $\mathcal{F}$ is larger than the sample size, then the covariance matrix of $\mathcal{F}$ is larger than $\mathcal{F}$. Therefore, in such a case, it is better to use the PCA formulation based on the data itself (PCA-form2).

One may reckon that using $[\mathcal{R}, \mathcal{M}]$ instead of $[\mathcal{F}, \mathcal{M}]$ may lead to loss of information due to dimension reduction and therefore lower the quality of imputation. However, as will be illustrated in the experiments, the differences between the mean squared error of the imputed version versus the ground truth for these approaches are only slightly different, and many times, PCAI seems to be slightly better. This is possibly because PCA retains the important information from the data while removing some noise, and therefore helps improving the imputation quality.
However, PCAI also has some shortcomings. For problems where the sample size $n$ is smaller than the number of features in the fully observed block $q$, if PCA-form1 is used, the covariance matrix has the size of $q\times q$, which is bigger than the size $n\times q$ of the fully observed partition $\mathcal{F}$. This may make the PCA dimension reduction process become computationally expensive, rendering PCAI to be slower than imputing directly on the original missing data. This issue will be illustrated in the experiment section.

\section{PCAI for classification (PIC)}\label{sec-pic}
In this section, we discuss a straightforward application of PCAI in classification, with a slight modification for classification problems where it is desirable to conduct a dimension reduction before training a model, such as when the number of features is much larger than the sample size.
To start, note that since PCAI conducts PCA on the fully observed partition $\mathcal{F}$, it reduces the dimensions for a portion of the data. Therefore, rather than imputing values using the PCAI framework and then conducting a dimension reduction step on $[\mathcal{F}, \mathcal{M}']$, one can perform dimension reduction on $\mathcal{M}'$ to get $\mathcal{R}'$, a PCA-reduced version of $\mathcal{M}'$. Then, one can use $[\mathcal{F}, \mathcal{R}']$ as reduced dimension data. As will be shown in the experiments, this speeds up the imputation and classification process significantly. This is the basic idea of our \textit{Principle component Imputation for Classification (PIC)} framework.

PIC operates as shown in Algorithm \ref{alg-pic}.
The procedure starts by performing PCA on the training fully observed partition $\mathcal{F}_{train}$, which gives the reduced version $\mathcal{R}_{train}$ of $\mathcal{F}_{train}$ and a projection matrix $V$. Next, we project $\mathcal{F}_{test}$ on $V$ to get the reduced version $\mathcal{R}_{test}$ of $\mathcal{F}_{test}$. 
Then, we impute $\mathcal{M}_{train}$ on $[\mathcal{R}_{train}, \mathcal{M}_{train}]$ to get the imputed version $\mathcal{M}_{train}'$. 
Next, we impute $\mathcal{M}_{test}$ on $[\mathcal{R}_{test}, \mathcal{M}_{test}]$ to get the imputed version $\mathcal{M}_{test}'$. 
After that, if $reduce_{miss}$ is set to true, the algorithm performs dimension reduction on $\mathcal{M}'_{train}, \mathcal{M}'_{test}$. 
Then, we train the classifier on $[\mathcal{R}_{train}, \mathcal{R}'_{train}]$, i.e., the union of the reduced version of $\mathcal{F}_{train}$ and the reduced version of $\mathcal{M}_{train}$. 
For prediction of a vector $\mathbf{x}\in \mathcal{D}$, we can decompose $\mathbf{x}$ into $\mathbf{x}=(\mathbf{x}_\mathcal{F}, \mathbf{x}_\mathcal{M})$. 
After that, we can project $\mathbf{x}_\mathcal{F}$ on $V$  to get a projection $\mathbf{r}$. 
Similarly, we can  project $\mathbf{x}_\mathcal{M}$ on $W$ to a get projection $\mathbf{r}'$. 
Finally, we can predict the label of $\mathbf{x}$ using the classifier $C$ with input $(\mathbf{r}, \mathbf{r}')$.  
\begin{algorithm}
\caption{PIC framework}\label{alg-pic}
\begin{algorithmic} 
\scriptsize
\REQUIRE

\STATE - $\mathcal{D} = [\mathcal{F}, \mathcal{M}]$ where $\mathcal{F}$ is the fully observed partition and $\mathcal{M}$ is the partition with missing values
\STATE - $reduce_{miss} = True/False$: if \emph{True}, perform dimension reduction on the imputed partitions; if \emph{False}, do not perform dimension reduction on the imputed partitions 
\STATE - $\mathcal{F}_{train}, \mathcal{F}_{test}$: the training and testing data of the fully observed partition $\mathcal{F}$, 
\STATE - $\mathcal{M}_{train}, \mathcal{M}_{test}$: the training and testing data of the partition that has missing data $\mathcal{M}$,
\STATE - Imputer $I$, classifier $C$, PCA algorithm $pca$

\hspace{-.4cm}\textbf{Procedure:}

\STATE $(\mathcal{R}_{train}, V) \leftarrow pca(\mathcal{F}_{train})$
\STATE $R_{test} \leftarrow \mathcal{F}_{test} V$
\STATE $\mathcal{M}_{train}'\leftarrow $ imputed version of $\mathcal{M}_{train}$ based on $[\mathcal{R}_{train},  \mathcal{M}_{train}]$
\STATE $\mathcal{M}_{test}'\leftarrow $ imputed version of $\mathcal{M}_{test}$ based on $[\mathcal{R}_{test}, \mathcal{M}_{test}]$

\IF{$\textbf{reduce}_{miss}$}
\STATE $(\mathcal{R}'_{train}, W) \leftarrow pca(\mathcal{M}'_{train})$
\STATE $\mathcal{R}'_{test} \leftarrow \mathcal{M}'_{test}W$
\STATE Train the classifier $C$ based on $[\mathcal{R}_{train}, \mathcal{R}'_{train}]$
\STATE Classify based on $[\mathcal{R}_{test}, \mathcal{R}'_{test}]$,

\ELSE
\STATE Train the classifier based on $[\mathcal{R}_{train}, \mathcal{M}'_{train}]$
\STATE  Classify based on $[\mathcal{R}_{test}, \mathcal{M}'_{test}]$
\ENDIF
\RETURN trained classifier $C$
\end{algorithmic}
\end{algorithm}

Note that when the number of features in the missing partition $\mathcal{M}$ is large, one may be interested in reducing the dimension of $\mathcal{M}'$, and therefore, set $reduce_{miss}$ to \textit{True}. However, when the number of features in the missing partition is small, one may want to keep it to \textit{False}. Also, since PIC is a straightforward application of PCAI for classification, the choice of PCA formulation should be used is similar to PCAI, which is analyzed in the previous section.
\section{Related Works}\label{sec-related}

Various works have been published on missing data imputation to deal with different situations, and with the rapid growth of data size \cite{NGUYEN2022108082}, various works have been done on PCA that are related to missing data, which mostly can be categorized into missing values imputation using PCA, or dimension reduction using PCA under missing values. 
Some typical works that make use of PCA for missing values imputation are probabilistic PCA for missing flow volume data imputation \cite{qu2009ppca}; chunk-wise iterative PCA for data imputation on datasets with many samples \cite{iodice2019single}; \cite{ilin2010practical} proposes a fast algorithm for PCA under missing data that help in case of sparse, high dimensional data;
and  \cite{audigier2016principal} proposed an imputation approach based on PCA and factorial analysis for mixed data. 
Next, PCA under missing values was first studied in \cite{dear1959principal}, where only one component and one imputation iteration are used. After that, \cite{andrews1997applications} proposes a method based on maximum likelihood PCA, where the method assigns large variance to missing values prior to implementing the method, which aim to guide the algorithm to fit a PCA model disregarding those points. Also, \cite{roweis1997algorithms} introduce EM algorithm for building a PCA model that can deal with missing data.
More recently, \cite{folch2015pca} proposes new techniques for building a PCA model with missing data. 
In addition, \cite{sportisse2020estimation} studies estimation and imputation in Probabilistic PCA when the data is missing not at random. 

Different from the previous approaches, PCAI is a framework to speed up the imputation process, which can be used with various imputation methods, including the aforementioned PCA imputation algorithms and the state-of-the-art imputation algorithms such as softImpute \cite{mazumder2010spectral}, MissForest \cite{stekhoven2012missforest}, GAIN \cite{yoon2018gain}. 
In addition, note that since PCAI and PIC conduct dimension reduction on the fully observed partition $\mathcal{F}$, and not the missing portion $\mathcal{M}$ if $reduce_{miss} = False$, they can handle missing data even if categorical features presents in the missing portion $\mathcal{M}$, when being used with imputers that's capable of handling categorical/mixed data (MissForest \cite{stekhoven2012missforest}, SICE \cite{khan2020sice}, FEMI \cite{rahman2016missing}, etc.). In addition, even if there exists categorical and continuous features in $\mathcal{M}$; or $reduce_{miss} = True$ and there exists categorical and continuous features in $\mathcal{M}$, one can easily adjust the algorithm to conduct PCA on continuous features only. The previously mentioned PCA based approaches are, however, can only be used for continuous data, because PCA requires the data to be continuous. 


\section{Experiments}\label{sec-exper}

\subsection{General experiment settings}

We compare the speed (seconds) and MSE of PCAI with  \textbf{direct imputation (DI)}, i.e., use an imputation algorithm directly on the dataset. The imputation approaches used for comparison: softImpute \cite{mazumder2010spectral,fancyimpute}, MissForest \cite{stekhoven2012missforest} \footnote{\url{https://pypi.org/project/missingpy/}}
and MICE \cite{buuren2010mice,scikit-learn}, KNNI, GAIN \cite{yoon2018gain} are implemented with default configurations. The codes will be available upon the acceptance of the paper. For PIC, we compare the five fold cross-validation (CV) score (accuracy, speed) of PIC when dimension reduction is applied on the imputed missing part (\textbf{PIC-reduce)}, when dimension reduction is not applied on the imputed missing part (\textbf{PIC}), and when PCA is applied to the imputed version on the full missing data (\textbf{DI-reduce}), and when no dimension reduction is applied to imputed data after direct imputation (\textbf{DI}). Here, the default PCA formulation is PCA-form1, unless specified otherwise. For all PCA computation, the number of eigenvectors is chosen so that the minimum amount of variance explained is 95\%. 

Details of the datasets used in the experiments are available in Table \ref{table_info_datasets}. 
All experiments are run on an AMD Ryzen 7 3700X CPU with 8 Cores, 16 processing threads, 3.6GHz, and 16GB RAM. 
We terminate an experiment if no result is produced after 6,500 seconds of running or if there arises a memory allocating issue, and we denote this as \textbf{NA} in the result tables. 

\begin{table}
\scriptsize
  \caption{Description of datasets used in our experiments}
  \label{table_info_datasets}
  \centering
  \begin{tabular}{lrrr}
    \toprule
    \cmidrule(r){1-2}
    Dataset     & \# Classes     & \# Features & \# Samples \\
    \midrule
    Parkinson \cite{sakar2019comparative} & 2 & 754 & 756 \\
    Fashion MNIST \cite{xiao2017fashion} & 10 & 784 & 70000 \\
    Gene \cite{Dua:2019} & 5 & 20531 & 801 \\
    \bottomrule
  \end{tabular}
\end{table}


\subsection{Performance of PCAI and PIC when the missing values in $\mathcal{M}$ are randomly simulated}\label{sec: pcai-pic}
\begin{table}[ht]
\scriptsize
\caption{(MSE, speed) for PCAI and DI on the Parkinson dataset with $q = 700$.} 
\label{tab-pcai-parkinson}
\centering
\begin{tabular}{@{\extracolsep{4pt}}llccc}
\toprule   
{} & {} & \multicolumn{3}{c}{missing rate} \\
 \cmidrule{3-5} 
 Imputer & Strategy & 20\% & 40\% & 60\%  \\ 
\midrule

\multirow{2}{*}{softImpute}
  & PCAI & (0.073, 0.860) & (0.185, 0.774) & (0.305, 0.875)    \\
  & DI & (0.072, 4.097) & (0.188, 4.043) & (0.308, 4.467)   \\
\addlinespace
  \multirow{2}{*}{MICE}
  & PCAI & (0.091, 139.811) & (0.186, 85.241) & (0.369, 109.815)  \\
  & DI & NA & NA & NA  \\
  
\addlinespace
  \multirow{2}{*}{GAIN}
  & PCAI & (0.254, 45.046) & (0.538, 43.938) & (0.779, 43.956)  \\
  & DI & (0.608, 69.839) & (1.097, 70.548) & (1.369, 70.293)   \\
  
\addlinespace
  \multirow{2}{*}{missForest}
  & PCAI & (0.064, 188.324) & (0.163, 178.849) & (0.292, 138.085)  \\
  & DI & (0.058, 905.002) & (0.160, 692.150) & (0.258, 449.415)   \\
  
\addlinespace
  \multirow{2}{*}{KNNI}
  & PCAI & (0.127, 0.355) & (0.299, 0.398) & (0.466, 0.416)  \\
  & DI & (0.113, 0.310) & (0.274, 0.337) & (0.426, 0.372)   \\

\bottomrule
\end{tabular}

\end{table}

\begin{table}[ht]
\scriptsize
\caption{(MSE, speed) for PCAI and DI on the Fashion MNIST dataset with $q = 700$. MissForest results all are NA, and therefore are removed from the tables.} 
\label{tab-pcai-fashion}
\centering
\begin{tabular}{@{\extracolsep{4pt}}llccc}
\toprule   
{} & {} & \multicolumn{3}{c}{missing rate} \\
 \cmidrule{3-5} 
 Imputer & Strategy & 20\% & 40\% & 60\%  \\ 
\midrule

\multirow{2}{*}{softImpute}
  & PCAI & (0.032, 22.408) & (0.066, 22.797) & (0.109, 25.603)    \\
  & DI & (0.032, 67.627) & (0.064, 69.349) & (0.107, 77.233)   \\
\addlinespace
  \multirow{2}{*}{MICE}
  & PCAI & (0.027, 2218.864) & (0.055, 1374.558) & (0.095, 1641.962)  \\
  & DI & NA & NA & NA  \\
  
\addlinespace
  \multirow{2}{*}{GAIN}
  & PCAI & (0.053, 65.730) & (0.091, 68.752) & (0.137, 69.743) \\
  & DI & (0.041, 97.898) & (0.079, 99.049) & (0.125, 96.317)   \\
  
\addlinespace
  
  \multirow{2}{*}{KNNI}
  & PCAI & (0.055, 1607.850) & (0.115, 2033.153) & (0.180, 2272.370)  \\
  & DI & (0.049, 3042.752) & (0.102, 3659.300) & (0.161, 3959.832)   \\

\bottomrule
\end{tabular}

\end{table}

Note that any datasets can be rearranged so that the first $q$ features are not missing and the remaining ones are missing. Therefore, without loss of generality, we assume that the first $q$ features of each dataset are not missing, and the remaining ones contain missing value(s). Then, we simulated missing data randomly on the missing partition $\mathcal{M}$ with missing rates 20\%, 40\%, and 60\%. Here, a missing rate of 20\% means that 20\% of the entries in the missing partition $\mathcal{M}$ are missing. The results for such experiments are reported in Tables \ref{tab-pcai-parkinson}, \ref{tab-pcai-fashion}, \ref{tab-pic-parkinson}. Due to space limit, the results related to PIC on Fashion MNIST are reported in the supplementary materials. 

\begin{table}[!ht]
\scriptsize
\caption{Five fold CV results (accuracy, speed) of SVM  on Parkinson with $q = 700$.} 
\label{tab-pic-parkinson}
\centering
\begin{tabular}{@{\extracolsep{4pt}}llccc}
\toprule   
{} & {} & \multicolumn{3}{c}{missing rate} \\
 \cmidrule{3-5} 
 Imputer & Strategy & 20\% & 40\% & 60\% \\ 
\midrule

\multirow{2}{*}{softImpute}
  & PIC-reduce & (0.862, 1.026) & (0.862, 1.137) & (0.862, 1.161)    \\
  & PIC & (0.858, 1.008) & (0.858, 1.079) & (0.859, 1.112)   \\
  & DI-reduce & (0.861, 4.116) & (0.862, 4.424) & (0.861, 4.718)   \\
  & DI & (0.858, 3.775) & (0.858, 3.912) & (0.855, 4.248)\\
\addlinespace
  \multirow{2}{*}{MICE}
  & PIC-reduce & (0.859, 204.605) & (0.861, 256.340) & (0.861, 240.211)    \\
  & PIC & (0.858, 524.739) & (0.859, 694.667) & (0.859, 925.426) \\
  & DI-reduce & NA & NA & NA  \\
  & DI & NA & NA & NA\\
\addlinespace
  \multirow{2}{*}{GAIN}
  & PIC-reduce & (0.857, 91.086) & (0.852, 102.861) & (0.848, 122.349)    \\
  & PIC & (0.851, 89.984) & (0.853, 104.773) & (0.853, 123.233)   \\
  & DI-reduce & (0.855, 130.349) & (0.851, 149.864) & (0.851, 181.135)   \\
  &DI & (0.846, 129.702) & (0.849, 152.031) & (0.852, 183.67)\\
\addlinespace
  \multirow{2}{*}{missForest}
  & PIC-reduce & (0.859, 204.850) & (0.861, 276.537) & (0.858, 153.783)   \\
  & PIC & (0.858, 202.939) & (0.861, 277.067) & (0.858, 153.463)   \\
  & DI-reduce & (0.861, 656.948) & (0.862, 729.872) & (0.861, 472.230)   \\
  &DI & (0.858, 655.750) & (0.861, 730.013) & (0.858, 472.388)\\
\addlinespace
  \multirow{2}{*}{KNNI}
  & PIC-reduce & (0.858, 0.533) & (0.861, 0.462) & (0.862, 0.625)    \\
  & PIC & (0.858, 0.513) & (0.861, 0.462) & (0.862, 0.607)   \\
  & DI-reduce & (0.862, 0.696) & (0.862, 0.642) & (0.859, 0.803)   \\
  &DI & (0.859, 0.438) & (0.859, 0.45) & (0.858, 0.552)\\
\bottomrule
\end{tabular}

\end{table}

From the tables, it is clear that the proposed frameworks reduce the imputation time significantly while maintaining competitive MSE/classification accuracy compared to DI, in most of the cases. For example, at the missing rate 20\% on the Parkinson dataset (Table \ref{tab-pic-parkinson}), when using GAIN for imputation, the running time of PIC-reduce (91.086s) is much lower compared to DI-reduce (130.349), the running time of PIC (89.984s) is also much lower compared to DI (129.702). Another example can be seen from Table \ref{tab-pcai-parkinson}, for the Parkinson dataset, at 20\% missing rate, when PCAI is applied to missForest, the running time reduces to 188.324s, which is almost $1/5$ of the DI (905.002s). Next, on Fashion MNIST (Table \ref{tab-pcai-fashion}), it is worth noticing that for MICE, DI cannot gives the results due to memory issue but PCAI can alleviate this issue and deliver the results.


%
For KNNI, the running time for KNNI between the PCAI approach and direct imputation for Parkinson (Table \ref{tab-pcai-parkinson}) is not much different. However, for the Fashion MNIST dataset, KNNI using the PCAI framework obviously deliver a competitive result in a significantly shorter time. Specifically, KNNI at a missing rate of 20\% on Fashion MNIST gives a result after only 1607.850 seconds, while DI takes up to 3,042.752 seconds. This is because Fashion MNIST (70000 samples) has much more samples than Parkinson (756 samples), and KNN need to do a lot of pairwise comparison. Therefore, PCAI and PIC would be extremely helpful for KNNI when the sample size and the number of fully observed features is large. Note that it does not require the number of features with missing data to be large or small.

From Table \ref{tab-pcai-parkinson}, we can see that PCAI  generates a lot of improvements in MSE for GAIN, in addition to improvements in speed. This is possibly because PCA reduces the number of features while the sample size remains the same, making such a deep learning approach more applicable to the newly reduced data. 

For PIC, 
the result for experiments related to PIC are provided in 
Table \ref{tab-pic-mnist}. From the tables, it is clear that PIC approaches reduce the imputation time significantly while maintaining competitive classification accuracy compared to PCA on $[\mathcal{F}, \mathcal{M}']$, in most of the cases.

\begin{table}[ht]
\caption{5 - fold cross validation results (accuracy, speed) of SVM for different imputation-classification strategies on the Fashion MNIST dataset with $q = 700$, when the missing values in $\mathcal{M}$ are simulated at random given rates.} 
\label{tab-pic-mnist}
\centering
\begin{tabular}{@{\extracolsep{4pt}}llccc}
\toprule   
{} & {} & \multicolumn{3}{c}{missing rate} \\
 \cmidrule{3-5} 
 Imputer & Strategy & 20\% & 40\% & 60\% \\ 
\midrule

\multirow{2}{*}{softImpute}
  & PIC-reduce & (0.891, 285.707) & (0.890, 279.263) & (0.889, 353.506)    \\
  & PIC & (0.891, 593.652) & (0.891, 598.726) & (0.890, 609.252)   \\
  & DI-reduce & (0.892, 467.406) & (0.891, 458.625) & (0.891, 486.162)   \\
  &DI & (0.892, 710.019) & (0.892, 644.585) & (0.891, 630.044)\\
\addlinespace
  \multirow{2}{*}{MICE}
  & PIC-reduce & (0.892, 2379.869) & (0.892, 1884.385) & (0.891, 2851.581)    \\
  & PIC & (0.892, 2416.485) & (0.891, 1907.529) & (0.892,  2897.839)    \\
  & DI-reduce & NA & NA & NA \\
  &DI & NA & NA & NA\\
\addlinespace
  \multirow{2}{*}{GAIN}
  & PIC-reduce & (0.892, 539.720) & (0.891, 580.754) & (0.891, 660.633)    \\
  & PIC & (0.891, 583.016) & (0.892, 608.384) & (0.891, 703.145)   \\
  & DI-reduce & (0.891, 1320.018) & (0.891, 1351.854) & (0.890, 1573.656)   \\
  &DI&(0.891, 846.002) & (0.892, 787.246) & (0.891, 863.685)\\
  
\addlinespace
  \multirow{2}{*}{KNNI}
  & PIC-reduce & (0.891, 3040.393) & (0.891, 4235.292) & (0.89, 6059.608)    \\
  & PIC & (0.891, 3110.041) & (0.891, 4279.305) & (0.889,  6088.542)   \\
  & DI-reduce &  NA
 & NA & NA  \\
&DI & NA & NA & NA\\
\bottomrule
\end{tabular}
\end{table}

\subsection{PIC under different PCA formulations and number of missing features}
The missing data in these experiments are generated at random as in Section \ref{sec: pcai-pic} and the five fold cross validation results of SVM on the Gene dataset with $q = 15000, 20000$, are shown in Table \ref{tab-pic-gene-15000} and Table \ref{tab-pic-gene-20000}. From these tables, one can see clearly that for datasets where the number of features are significantly higher than the number of samples such as Gene, PCA-form2, which is based on the input data ($\mathcal{F}$ specifically) gives much faster computations compared to PCA-form1, and also is faster than direct imputation-classification without PCA. In addition, when PCA-form1 is used, even though PIC and PIC-reduce  are faster than PCA on directly imputed data (DI-reduce), they are still much slower than direct imputation - classification without PCA. 

Interestingly, the accuracy PIC and PIC-reduce are almost identical to PCA on directly imputed data, and are higher than direct imputation - classification without PCA. Next, note that the main idea of the proposed methods is to reduce the dimension of the $\mathcal{F}$ to speed up the imputation. Therefore, we have made no assumption about the number of features in the missing portion $\mathcal{M}$. In Table \ref{tab-pic-gene-15000} and Table \ref{tab-pic-gene-20000}, $q= 15000, 20000$, which means 5,531 and 531 missing features in $\mathcal{M}$, respectively. This means PIC can handle datasets where $\mathcal{M}$ has many features.
\begin{table}[!h]
\scriptsize
\caption{Five fold CV results (accuracy, speed) of SVM for softImpute based strategies on the Gene dataset when $q = 15000$.} 
\label{tab-pic-gene-15000}
\centering
\begin{tabular}{@{\extracolsep{4pt}}llccc}
\toprule   
{} & {} & \multicolumn{3}{c}{missing rate} \\
 \cmidrule{3-5} 
  & Strategy & 20\% & 40\% & 60\% \\ 
\midrule

\multirow{2}{*}{PCA-form1}
  & PIC-reduce & (0.994, 2250.451)   & (0.992, 2412.082)  & (0.992, 2415.434)     \\
  & PIC &  (0.992, 2429.114) & (0.992, 2276.354)  & (0.992, 2284.414)   \\
  & DI-reduce  & (0.994, 5018.368) &  (0.994, 4529.766) &   (0.994, 3785.947) \\ 
\addlinespace
  \multirow{2}{*}{PCA-form2}
  & PIC-reduce & (0.995, 47.850)  & (0.992, 56.638) &  (0.992, 63.980)    \\
  & PIC & (0.992, 48.356) & (0.992, 57.113) & (0.992, 64.786) \\
  & DI-reduce & (0.995, 99.698) & (0.992, 110.993) & (0.994,128.083)  \\
\addlinespace
 No PCA & DI & (0.985, 71.884)  & (0.985, 74.812) &  (0.985, 92.309)    \\
 \bottomrule
\end{tabular}
\end{table}

{\setlength\intextsep{1.5pt}
\begin{table}[!ht]
\scriptsize
\caption{Five fold CV results (accuracy, speed) of SVM for softImpute based strategies on the Gene dataset when $q=20000$.} 
\label{tab-pic-gene-20000}
\centering
\begin{tabular}{@{\extracolsep{4pt}}llccc}
\toprule   
{} & {} & \multicolumn{3}{c}{missing rate} \\
 \cmidrule{3-5} 
  & Strategy & 20\% & 40\% & 60\% \\ 
\midrule

\multirow{2}{*}{PCA-form1}
      & PIC-reduce & (0.994, 2578.910)  & (0.994, 4001.717)  &(0.994, 3848.950) \\
  & PIC &  (0.994, 2583.717)& (0.994, 4144.157)  &  (0.994, 4057.188)  \\
  & DI-reduce & (0.995, 2891.994) &  (0.994, 4476.563) &   (0.995, 4332.869) \\
\addlinespace
  \multirow{2}{*}{PCA-form2}
  & PIC-reduce & (0.995, 8.666)  & (0.995, 9.597) &  (0.995, 10.639)    \\
  & PIC & (0.995, 8.566) & (0.995, 9.464) & (0.995, 10.491) \\
  & DI-reduce & (0.995, 87.298) & (0.995, 78.814) & (0.995, 79.843)  \\
\addlinespace
 No PCA & DI & (0.985, 74.06)  & (0.985, 71.6) &  (0.985, 84.963)    \\
 \bottomrule
\end{tabular}
\end{table}
}

\section{Conclusion and Remarks}\label{sec-concl}
We have presented two novel frameworks for datasets where many continuous features are fully observed, PCAI and PIC, that can speed up imputation algorithms significantly while having competitive accuracy MSE/accuracy compared to direct imputation and alleviate the memory issue for some imputation approaches such as MICE, kNN. In addition, the frameworks can be used even when some or all of the missing features are categorical or when the number of missing features is large. Note that when the sample size is significantly larger than the number of fully observed features, PCA-form1 should be used since, in such a case, the covariance matrix is much smaller than $\mathcal{F}$, making it faster than PCA-form2. On the other hand, when the number of fully observed features is significantly larger than the sample size, PCA-form2 should be preferred, as the covariance matrix is bigger than $\mathcal{F}$ itself in such a case. A limitation of the proposed framework is that if there are not many fully observed continuous features, then due to the computational cost of PCA, the proposed frameworks may not lead to any improvement in speed. 

Even though PIC is only introduced for classification, the same strategy can be applied to a regression problem. We would like to explore that in the future. Moreover, since various dimension reduction techniques such as sparse PCA \cite{jenatton2010structured}, incremental PCA \cite{ross2008incremental}, truncated SVD \cite{halko2011finding} have been developed to suit different scenarios, it is worth investigating different dimension reduction techniques for PCAI and PIC. In addition, it would be interesting to explore if applying a PCA variant to the missing partition $\mathcal{M}$ would result in even a more efficient method for datasets with continuous features in the missing partition.

%
%
\small

\bibliographystyle{splncs04}
\bibliography{bibfile}

%


\appendix

\end{document}


\maketitle

\appendix

In the following, we present the supplementary materials for the paper \textit{Principal Components Analysis based frameworks for efficient missing data imputation algorithms}.
\section{PIC on the Fashion MNIST dataset}

\begin{table}[ht]
\caption{5 - fold cross validation results (accuracy, speed) of SVM for different imputation-classification strategies on the Fashion MNIST dataset with $q = 700$, when the missing values in $\mathcal{M}$ are simulated at random given rates.} 
\label{tab-pic-mnist}
\centering
\begin{tabular}{@{\extracolsep{4pt}}llccc}
\toprule   
{} & {} & \multicolumn{3}{c}{missing rate} \\
 \cmidrule{3-5} 
 Imputer & Strategy & 20\% & 40\% & 60\% \\ 
\midrule

\multirow{2}{*}{softImpute}
  & PIC-reduce & (0.891, 285.707) & (0.890, 279.263) & (0.889, 353.506)    \\
  & PIC & (0.891, 593.652) & (0.891, 598.726) & (0.890, 609.252)   \\
  & DI-reduce & (0.892, 467.406) & (0.891, 458.625) & (0.891, 486.162)   \\
  &DI & (0.892, 710.019) & (0.892, 644.585) & (0.891, 630.044)\\
\addlinespace
  \multirow{2}{*}{MICE}
  & PIC-reduce & (0.892, 2379.869) & (0.892, 1884.385) & (0.891, 2851.581)    \\
  & PIC & (0.892, 2416.485) & (0.891, 1907.529) & (0.892,  2897.839)    \\
  & DI-reduce & NA & NA & NA \\
  &DI & NA & NA & NA\\
\addlinespace
  \multirow{2}{*}{GAIN}
  & PIC-reduce & (0.892, 539.720) & (0.891, 580.754) & (0.891, 660.633)    \\
  & PIC & (0.891, 583.016) & (0.892, 608.384) & (0.891, 703.145)   \\
  & DI-reduce & (0.891, 1320.018) & (0.891, 1351.854) & (0.890, 1573.656)   \\
  &DI&(0.891, 846.002) & (0.892, 787.246) & (0.891, 863.685)\\
\addlinespace
  \multirow{2}{*}{missForest}
  & PIC-reduce & NA & NA & NA   \\
  & PIC & NA & NA & NA   \\
  & DI-reduce & NA & NA & NA   \\
  
\addlinespace
  \multirow{2}{*}{KNNI}
  & PIC-reduce & (0.891, 3040.393) & (0.891, 4235.292) & (0.89, 6059.608)    \\
  & PIC & (0.891, 3110.041) & (0.891, 4279.305) & (0.889,  6088.542)   \\
  & DI-reduce &  NA
 & NA & NA  \\
&DI & NA & NA & NA\\
\bottomrule
\end{tabular}
\end{table}

\section{PIC on Fashion MNIST under monotone missing data}
Table \ref{tab-pic-mnist-monotone-full} shows the full results for PIC on Fashion MNIST under monotone missing data, which corresponds to the experiment settings in \textit{Section 7.3: Performance on nonrandomly missing data} in the paper.  

Interestingly, in this case, for KNNI, at 20\% missing rates, DI-reduce is faster than PIC-reduce, and DI is faster than DI. However, as the missing rates increase to 40\% or 60\%, PIC-reduce is significantly faster than PIC, and PIC is also significantly faster than DI. Also, note that in Table \ref{tab-pic-mnist}, when KNNI is used, DI and DI-reduce gives NA (due to memory issue). This shows that how the data is missing affects the performance of KNNI, and that PCAI and PIC have the ability to alleviate the memory requirements of KNNI.

\begin{table}[!ht]
\caption{Five fold CV results (accuracy, speed) of SVM on monotone data generated on Fashion MNIST. MICE and MissForest are excluded from the table because all the corresponding results are NA.} 
\label{tab-pic-mnist-monotone-full}
\centering
\begin{tabular}{@{\extracolsep{4pt}}llccc}
\toprule   
{} & {} & \multicolumn{3}{c}{missing rate} \\
 \cmidrule{3-5} 
 Imputer & Strategy & 20\% & 40\% & 60\% \\ 
\midrule

\multirow{2}{*}{softImpute}
  & PIC-reduce & (0.889, 409.676) & (0.889, 421.671) &(0.889, 369.49)  \\
  & PIC & (0.889, 507.626) & (0.89, 543.309) & (0.889, 480.452)   \\
  & DI-reduce & (0.89, 439.268) & (0.89, 528.892) & (0.889, 395.646)  \\
  & DI & (0.891, 738.494) & (0.891, 872.616) & (0.89, 646.781)\\
\addlinespace
  \multirow{2}{*}{GAIN}
  & PIC-reduce & (0.886, 462.478) & (0.883, 429.173) & (0.882, 449.786)    \\
  & PIC & (0.886, 493.399) & (0.882, 484.232) & (0.881, 496.066)  \\
  & DI-reduce & (0.891, 543.803) & (0.89, 431.947) & (0.89, 454.981)  \\
  &DI & (0.892, 902.049) & (0.891, 754.794) & (0.891, 780.686)\\
\addlinespace
  \multirow{2}{*}{KNNI}
  & PIC-reduce & (0.89, 2220.308)& (0.89, 2344.1) & (0.889, 2800.795)  \\
  & PIC & (0.89, 2278.491) & (0.891, 2381.274) & (0.89, 2838.674)\\
  & DI-reduce & (0.89, 1720.832) & (0.89, 2614.603) &  (0.89, 3173.561) \\
  &DI & (0.891, 1958.2) & (0.891, 2854.299) & (0.89, 3416.44)\\
\bottomrule
\end{tabular}
\end{table}

\section{PCAI on Fashion MNIST under monotone missing data}
Table \ref{tab-pcai-fashion-monotone} shows the full results for PCAI on Fashion MNIST under monotone missing data, which corresponds to the experiment settings in \textit{Section 7.3: Performance on nonrandomly missing data} in the paper. 

\begin{table}[ht]
\caption{(MSE, speed) for PCAI and DI on the Fashion MNIST dataset under monotone data. MICE and MissForest are excluded from the table because all the corresponding results are NA.} 
\label{tab-pcai-fashion-monotone}
\centering
\begin{tabular}{@{\extracolsep{4pt}}llccc}
\toprule   
{} & {} & \multicolumn{3}{c}{missing rate} \\
 \cmidrule{3-5} 
 Imputer & Strategy & 20\% & 40\% & 60\%  \\ 
\midrule

\multirow{2}{*}{softImpute}
  & PCAI & (0.052, 50.057)& (0.098, 61.523) & (0.014, 63.632)    \\
  & DI & (0.050, 170.488) & (0.095, 212.620) & (0.013, 212.620)   \\
  
\addlinespace
  \multirow{2}{*}{GAIN}
  & PCAI &  (0.34, 67.052) & (0.793, 68.788)&(0.96, 67.247) \\
  & DI & (0.367, 108.869) & (0.898, 106.761) & (1.424, 109.571)     \\
  
\addlinespace
  
  \multirow{2}{*}{KNNI}
  & PCAI & (0.054, 1210.02) & (0.099, 1929.294) & (0.139, 2349.668)  \\
  & DI & (0.046, 2199.876) & (0.085, 3507.741) & (0.120, 4181.563)    \\

\bottomrule
\end{tabular}

\end{table}